\documentclass[letterpaper]{article} 
\usepackage{aaai25}   
\usepackage{times}  
\usepackage{helvet}  
\usepackage{courier}  
\usepackage[hyphens]{url}  
\usepackage{graphicx} 
\usepackage{multirow}
\usepackage{booktabs}
\usepackage{colortbl}
\usepackage{bibentry}
\usepackage[dvipsnames]{xcolor}
\usepackage{subfigure}
\usepackage{natbib}  
\usepackage{caption} 
\usepackage{algorithm}
\usepackage{algorithmic}
\usepackage{mdframed}
\usepackage{listings}
\usepackage[many]{tcolorbox}
\tcbset{
   promptbox/.style={
       enhanced,
       colback=white,
       colframe=black!50,
       fonttitle=\bfseries,
       boxrule=0.5pt,
       colbacktitle=gray!10,
       coltitle=black, 
      left=1pt,   
      right=1pt,  
      top=1pt,      
      bottom=1pt, 
      width=\columnwidth,
   }
}
\usepackage{newfloat}
\usepackage{listings}
\DeclareCaptionStyle{ruled}{labelfont=normalfont,labelsep=colon,strut=off} 
\floatstyle{ruled}
\newfloat{listing}{tb}{lst}{}
\floatname{listing}{Listing}
\title{Improving Fine-grained Visual Understanding in VLMs \\through Text-Only Training}

\author {
    Dasol Choi\textsuperscript{\rm 1,\rm 5},
    Guijin Son\textsuperscript{\rm 1},
    Soo Yong Kim\textsuperscript{\rm 2},
    Gio Paik\textsuperscript{\rm 3},
    Seunghyeok Hong\textsuperscript{\rm 4}\thanks{Corresponding Author}
}
\affiliations {
    \textsuperscript{\rm 1}Yonsei University,
    \textsuperscript{\rm 2}Seoul National University,
    \textsuperscript{\rm 3}Sejong University,
    \textsuperscript{\rm 4}The University of Suwon,
        \textsuperscript{\rm 5}MODULABS\\
    \{dasolchoi, guijin.son\}@yonsei.ac.kr, ksyint1111@snu.ac.kr,
    \{giopaik, shongdr\}@gmail.com
}

\begin{document}

\maketitle

\begin{abstract}
Visual-Language Models (VLMs) have become a powerful tool for bridging the gap between visual and linguistic understanding. However, the conventional learning approaches for VLMs often suffer from limitations, such as the high resource requirements of collecting and training image-text paired data. Recent research has suggested that language understanding plays a crucial role in the performance of VLMs, potentially indicating that text-only training could be a viable approach. In this work, we investigate the feasibility of enhancing fine-grained visual understanding in VLMs through text-only training. Inspired by how humans develop visual concept understanding, where rich textual descriptions can guide visual recognition, we hypothesize that VLMs can also benefit from leveraging text-based representations to improve their visual recognition abilities. We conduct comprehensive experiments on two distinct domains: fine-grained species classification and cultural visual understanding tasks. Our findings demonstrate that text-only training can be comparable to conventional image-text training while significantly reducing computational costs. This suggests a more efficient and cost-effective pathway for advancing VLM capabilities, particularly valuable in resource-constrained environments.
\end{abstract}

\begin{figure*}[t]
\centering
\includegraphics[width=0.90\linewidth]{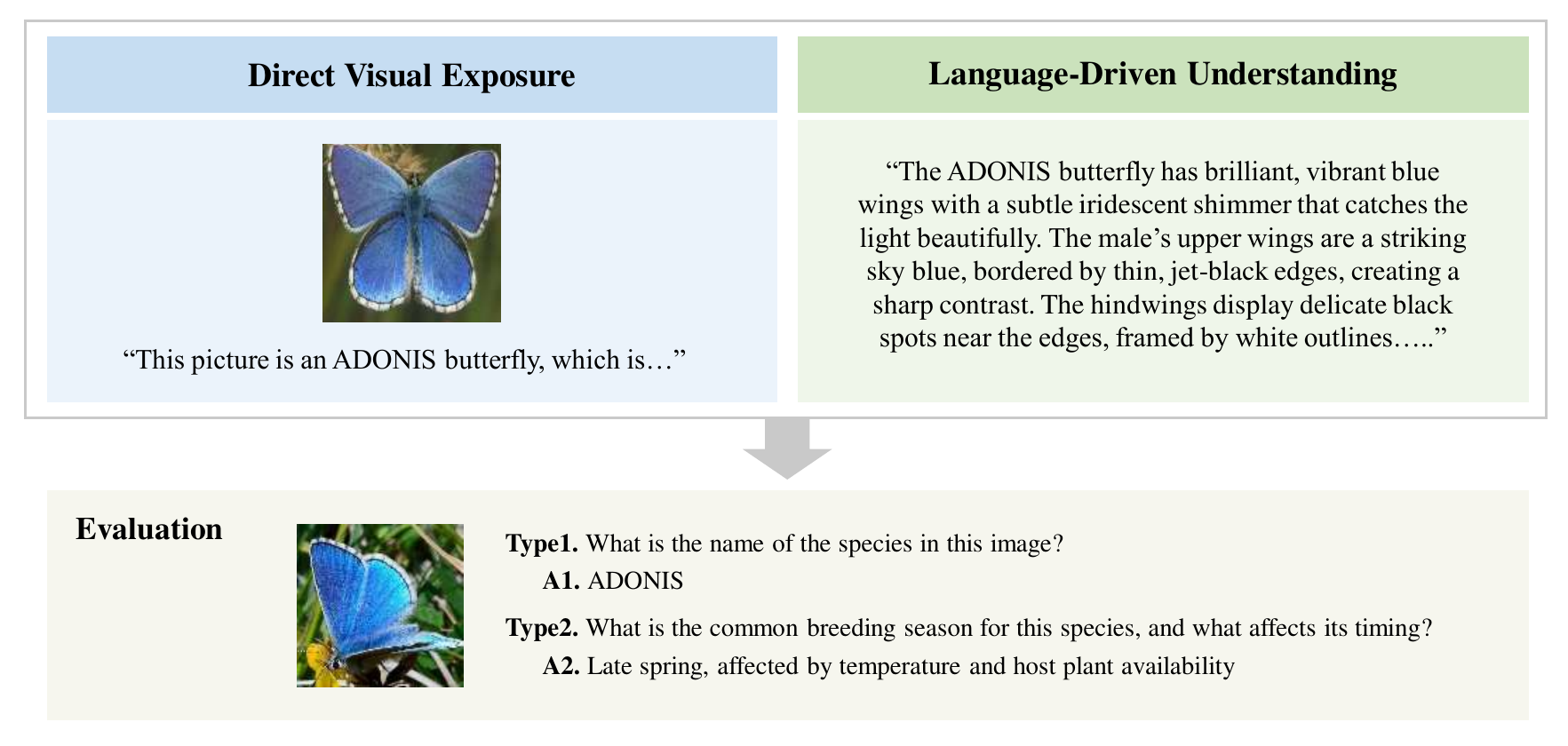}
\vspace{-0.6em}
\caption{\footnotesize{Illustration of our approach comparing direct visual exposure and language-driven understanding. Left: Traditional image-text paired training with direct visual exposure. Right: Our text-only training approach using detailed descriptions. Bottom: Example evaluation setting where both approaches are tested on the same visual understanding tasks.}}
\label{fig:concept}
\end{figure*}

\section{Introduction}

Recent advances in Vision-Language Models (VLMs) have revolutionized the way artificial intelligence systems understand and process visual information~\cite{clip, gpt4v, llava, InstructBLIP}. These models have achieved remarkable success across various tasks, from basic image captioning to complex visual reasoning, by effectively combining visual and linguistic representations~\cite{Palm-e, flamingo, blip}. However, the current paradigm of VLM training faces significant challenges: it requires extensive collections of image-text paired data and demands substantial computational resources for visual processing~\cite{zhang2024vision, LAION-5B, patterson2021carbon}.

The conventional wisdom suggests that visual understanding necessitates direct exposure to images. However, emerging research is challenging this assumption, revealing that language understanding often plays a more fundamental role in VLMs' performance than previously recognized. \citet{LENS} demonstrated through their LENS framework that VLMs heavily rely on their language understanding capabilities, showing that decomposing visual inputs into detailed textual descriptions can achieve comparable or better performance than end-to-end vision-language models. Similarly, \citet{Lovm} revealed that language-only evaluation could effectively predict zero-shot performance without access to visual data, suggesting that textual representations can sometimes dominate over visual features in certain tasks~\citep{caron2024webscalevisualentityrecognition}. These findings raise an intriguing possibility: could we enhance VLMs' visual understanding capabilities through text-only training?

Our approach draws inspiration from how humans develop visual concept understanding. Consider how children learn to recognize and categorize visual entities: in early stages, young children primarily learn through direct visual exposure~\cite{carey1999sources}. For instance, they learn about the ``sea" by visiting beaches or looking at pictures, building their recognition abilities from direct experiences. However, as their cognitive and linguistic abilities develop during their early school years, this learning process evolves~\cite{gentner2010mutual, Waxman2007EarlyWA}. Children become increasingly capable of understanding new visual concepts through textual descriptions alone. A description such as ``the sea is a vast expanse of water with a sky above and sand below" can enable them to form accurate mental representations, demonstrating how well-structured language can guide visual understanding effectively. This natural progression from pure visual learning to language-guided visual understanding suggests that detailed linguistic descriptions can effectively facilitate visual recognition abilities. We hypothesize that VLMs, similar to this human cognitive development, can leverage rich textual descriptions to enhance their visual recognition capabilities while overcoming the limitations of traditional image-and-text training approaches.

To test this hypothesis, we conduct experiments on two domains: a fine-grained species classification task of butterfly species and a cultural visual understanding task using a Korean cultural dataset. Our results demonstrate that text-only training can significantly enhance VLMs' visual recognition capabilities while substantially reducing computational demands. This approach offers a more efficient and cost-effective pathway for advancing VLM capabilities, which is particularly valuable in scenarios where computational resources or image-text paired data are limited.

\section{Methodology}

\subsection{Datasets}
We select two datasets based on two key criteria: (1) domain specificity requiring fine-grained visual understanding and (2) challenging aspects that current VLMs struggle with. The chosen datasets are the Butterflies and Moths dataset~\cite{butterfly}\footnote{hereafter referred to as \textsc{Butterfly} dataset}, which contains rich visual elements for species classification, and the Korean Cultural Understanding VQA Dataset~\cite{k-viscuit} (\textsc{K-viscuit}), which requires contextual and cultural visual understanding.

\subsubsection{Training Dataset}

The training datasets are created in two versions for each domain: one with image-text pairs and one with text-only. The image-text version contains one image per keyword along with a text description, while the text-only version contains textual descriptions without any images. This setup allows us to directly compare the learning outcomes of image-text and text-only training.

For the \textsc{Butterfly} training data, we randomly sample one image per species from the training set. We then prompt GPT-4o~\cite{GPT-4o} to generate textual descriptions, including visual details (e.g., color patterns, wing shapes) and biological characteristics, such as habitat and behavior. The \textsc{K-viscuit} dataset, however, only provides a test set without materials that may be leveraged for training. Accordingly, we collect one image from the internet per keyword. Following the collection, we generate textual descriptions via GPT-4o. The training datasets for \textsc{Butterfly} and \textsc{K-viscuit} consist of 100 and 237 samples, correspondingly. Detailed prompts and generated examples for both datasets are available in Appendix B.

\subsubsection{Evaluation Datasets}

For the \textsc{Butterfly} dataset, we create two types of evaluation settings (400 questions total) to assess VLM performance. Type 1 is a multiple-choice VQA where the model needs to identify the butterfly species from the image (200 questions). We combine correct image-specie pairs with incorrect options randomly selected from a list of butterfly species to make the question. Type 2 is a more complex multiple-choice VQA where the model needs to answer questions about the visual, ecological, and biological characteristics of the butterfly (200 questions). Using images taken from the test dataset, the questions are generated via the GPT-4o model. Examples of Type 2 questions are provided in Appendix B. Similarly, the \textsc{K-viscuit} benchmark originally consisted of two subsets. Type 1 questions focus on visual recognition (237 questions), while Type 2 questions involve more complex reasoning about cultural context, such as understanding the historical significance or usage of traditional objects (420 questions).

\begin{table*}[t]
\centering
\fontsize{9}{10}\selectfont
\renewcommand\arraystretch{1.1}
\begin{tabular}{c|c|c|ccc}
\toprule
\textbf{Model} & \textbf{Dataset} & \textbf{Type} & \textbf{Original} & \textbf{Image+Text} & \textbf{Text-only} \\
\midrule
\multirow{6}{*}{LLaVA-1.6-7B}
& \multirow{3}{*}{Butterfly} & Type 1 & 28.00 & 30.50 (\textcolor{green}{+2.50}) & 30.50 (\textcolor{green}{+2.50}) \\
& & Type 2 & 47.00 & 55.00 (\textcolor{green}{+8.00}) & 54.50 (\textcolor{green}{+7.50}) \\
& & Total & 37.50 & 42.75 (\textcolor{green}{+5.25}) & 42.50 (\textcolor{green}{+5.00}) \\
\cmidrule{2-6}
& \multirow{3}{*}{K-viscuit} & Type 1 & 44.30 & 56.96 (\textcolor{green}{+12.66}) & 51.05 (\textcolor{green}{+6.75}) \\
& & Type 2 & 56.90 & 61.67 (\textcolor{green}{+4.77}) & 59.05 (\textcolor{green}{+2.15}) \\
& & Total & 52.36 & 59.97 (\textcolor{green}{+7.61}) & 56.16 (\textcolor{green}{+3.80}) \\
\midrule
\multirow{6}{*}{Qwen2-VL-7B}
& \multirow{3}{*}{Butterfly} & Type 1 & 75.00 & 76.50 (\textcolor{green}{+1.50}) & 78.00 (\textcolor{green}{+3.00}) \\
& & Type 2 & 60.50 & 60.00 (-0.50) & 60.50 (+0.00) \\
& & Total & 67.75 & 68.25 (\textcolor{green}{+0.50}) & 69.25 (\textcolor{green}{+1.50}) \\
\cmidrule{2-6}
& \multirow{3}{*}{K-viscuit} & Type 1 & 64.14 & 71.31 (\textcolor{green}{+7.17}) & 74.26 (\textcolor{green}{+10.12}) \\
& & Type 2 & 67.86 & 70.95 (\textcolor{green}{+3.09}) & 69.76 (\textcolor{green}{+1.90}) \\
& & Total & 66.51 & 71.08 (\textcolor{green}{+4.57}) & 71.39 (\textcolor{green}{+4.88}) \\
\bottomrule
\end{tabular}
\vspace{-0.6em}
\caption{Performance comparison across models and training approaches (accuracy \%). The table includes a detailed breakdown by question types: Type 1 questions focus on visual recognition tasks, while Type 2 questions involve more complex reasoning. Values in parentheses indicate performance gains compared to the Original performance.}\label{tab:main_results}
\end{table*}

\subsection{Models}
For our experiments, we employ 7B VLMs from two family of models: Qwen2-VL~\cite{Qwen2-VL} and LLaVA-1.6~\cite{llave-next}. Our model selection is primarily driven by three factors: (1) the open-source nature, allowing for full fine-tuning experiments; (2) computational feasibility with our available resources; and (3) their prominence in the VLM community. Considering the relatively small size of the training dataset, each with fewer than 500 samples, we find it insufficient for training larger models. In our preliminary experiments, we observe training instability starting from 13B. Large-scale experiments with bigger datasets will be conducted in future research. Detailed model architectures and configurations are provided in Appendix A.



\section{Experimental Results}

\subsection{Main Results}
Our experimental results demonstrate that text-only training achieves comparable performance to conventional image-text training. As shown in Table~\ref{tab:main_results}, text-only training improves model performance across most evaluated criteria. The only exception is Type 2 of the \textsc{Butterfly} dataset. However, even in this case, while Image+Text training exhibits negative changes, text-only training does not.

For models with relatively lower initial performance like LLaVA-1.6-7B (Butterfly: 37.50\%), both training approaches show substantial improvements, with text-only training achieving a +5.00\%p gain compared to image-text training's +5.25\%p. In contrast, Qwen2-VL-7B, which starts with higher baseline performance (\textsc{Butterfly}: 67.75\%), shows more modest but still positive gains, with text-only training actually achieving slightly better improvement (+1.05\%p) compared to image-text training (+0.50\%p).

The effectiveness of text-only training also varies across datasets. In the \textsc{K-viscuit} dataset, which involves cultural and contextual understanding, text-only training demonstrates particularly strong performance. Notably, with Qwen2-VL-7B, text-only training slightly outperforms image-text training (71.39\% vs. 71.08\%), suggesting that detailed textual descriptions may be especially effective for conveying cultural visual concepts. 

\begin{figure*}[t]
\centering
\subfigure[Training Time]{
    \includegraphics[width=0.32\textwidth]{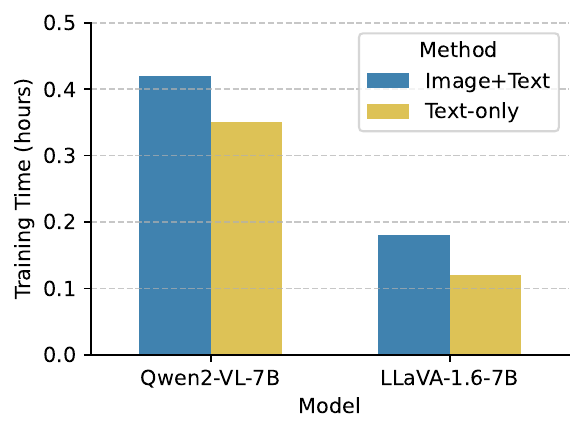}
}
\subfigure[Peak Memory Usage]{
    \includegraphics[width=0.32\textwidth]{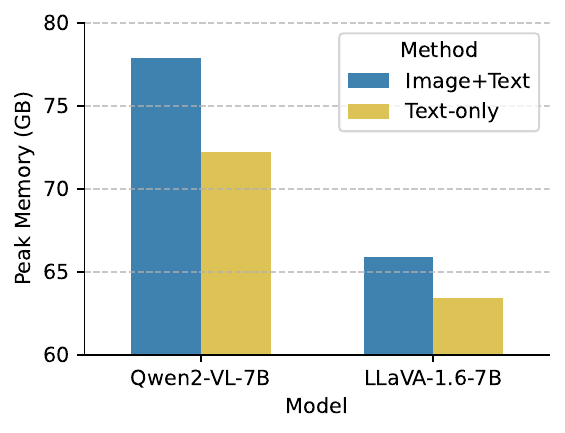}
}
\subfigure[Energy Consumption]{
    \includegraphics[width=0.32\textwidth]{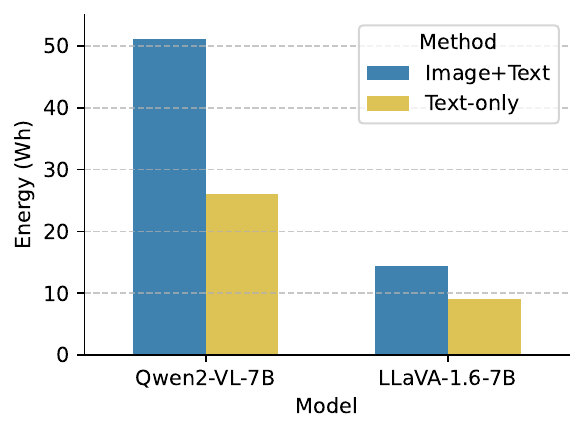}
}
\vspace{-0.6em}
\caption{Resource efficiency comparison between Image+Text and Text-only training approaches.}
\label{fig:resource_efficiency}
\end{figure*}

\subsection{Analysis by Question Types}

As detailed in Table~\ref{tab:main_results}, the impact of text-only training shows distinct patterns across different types of questions, providing insights into its strengths and limitations.

\subsubsection{Visual Recognition Tasks (Type 1)}

In visual recognition questions, text-only training demonstrates competitive and sometimes superior performance, particularly in tasks involving cultural understanding. For instance, on the \textsc{K-viscuit} dataset with Qwen2-VL-7B, text-only training achieves 74.26\% accuracy compared to 71.31\% with image-text training. This suggests that well-structured textual descriptions can effectively capture and convey visual features, even without direct image exposure during training.

For the \textsc{Butterfly} dataset, despite the task's requirement for fine-grained visual distinction, both approaches show comparable performance in Type 1 questions (77.50\% vs 78.50\% with Qwen2-VL-7B). This indicates that detailed textual descriptions can successfully capture subtle visual differences between species, making them as effective as image-based training for certain recognition tasks.

\subsubsection{Complex Reasoning Tasks (Type 2)}

Text-only training demonstrates particularly interesting results in complex reasoning tasks, suggesting its potential for higher-order visual understanding. In the \textsc{Butterfly} dataset with Qwen2-VL-7B, text-only training slightly outperforms image-text training (50.50\% vs 50.00\%). This indicates that textual descriptions may provide cleaner signals for learning complex visual concepts compared to potentially noisy image features.

In the \textsc{K-viscuit dataset}, text-only training maintains competitive performance in Type 2 questions (69.76\% vs 70.95\% with Qwen2-VL-7B), demonstrating that cultural and contextual understanding can be effectively learned through textual descriptions alone. This finding is particularly significant as it suggests that text-only training can support both basic visual recognition and more complex reasoning tasks effectively.

\subsection{Evidence Against Data Contamination}


One may be tempted to conclude that the benefits of text-only training are to test set contamination, where the model leverages patterns in the text descriptions to infer relationships between the questions and options rather than genuinely enhancing its image understanding capabilities. Accordingly, in this section, we conduct image-free evaluations on both the original and text-only trained models to verify that our model's improved performance is not due to simple memorization or data contamination. 

\begin{table}[ht]
\centering
\renewcommand\arraystretch{1.1}
\fontsize{9}{11}\selectfont
\begin{tabular}{lccc}
\toprule
\textbf{Setting} & \textbf{Overall} & \textbf{Type 1} & \textbf{Type 2} \\
\midrule
 \multicolumn{4}{l}{\textit{Model: LLaVA-1.6}}\\
Original (w/ image) & 52.36 & 44.30 & 56.90 \\
Original (no image) & 36.83 & 32.91 & 39.05 \\
Text-only (w/ image) & 56.16 & 51.05 & 59.05 \\
Text-only (no image) & 42.77 & 40.51 & 44.05 \\
\midrule
 \multicolumn{4}{l}{\textit{Model: Qwen2-VL}}\\
Original (w/ image) & 66.51 & 64.14 & 67.86 \\
Original (no image) & 45.97 & 45.99 & 45.95 \\
Text-only (w/ image) & 71.39 & 74.26 & 69.76 \\
Text-only (no image) & 47.18 & 47.68 & 46.90 \\
\bottomrule
\end{tabular}
\caption{\footnotesize Performance on the \textsc{K-viscuit} dataset with and without image inputs.}
\label{tab:contamination_analysis}
\end{table}

The experimental results provide strong evidence against contamination:

\begin{itemize}
   \item Both models show consistent patterns of performance degradation when images are removed (LLaVA: \-15.5\%, Qwen2-VL: \-20.5\%)
   \item The performance drop in Type 1 tasks (visual recognition) is substantial (LLaVA: 51.05\% → 40.51\%, Qwen2-VL: 74.26\% → 47.68\%), as expected for tasks requiring direct visual understanding. Importantly, Type 2 tasks (cultural understanding) also show significant degradation, indicating the models learn meaningful visual-semantic connections rather than superficial patterns
   \item Text-only training improves performance while maintaining the characteristic performance drops without images (13.4\% for LLaVA, 24.2\% for Qwen2-VL), demonstrating that the improvements stem from enhanced visual-linguistic alignment rather than text memorization
\end{itemize}

Table~\ref{tab:contamination_analysis} shows the performance comparison between models with and without image inputs. If the performance gains were due to contamination or mere memorization of text patterns, we would expect similar performance levels with and without images, and the pattern would likely vary between models. Instead, we observe consistent and substantial performance degradation across different architectures and task types; hence, text-only training is not a result of superficial pattern matching or data contamination.

\subsection{Resource Comparison}
Beyond performance improvements, text-only training demonstrates significant advantages in computational efficiency. As shown in Figure~\ref{fig:resource_efficiency}, we analyze three key metrics: training time, GPU memory usage, and energy consumption. 

In terms of training time, text-only training shows notable reductions for both models. 
The training time decreases by 33.3\% and 16.7\% for LLaVA-1.6-7B and Qwen2-VL-7B, respectively. This efficiency gain is primarily attributed to the elimination of image processing overhead. Peak GPU memory usage also benefits from text-only training. LLaVA-1.6-7B shows a reduction from 65.87GB to 63.44GB, while Qwen2-VL-7B demonstrates a more substantial decrease from 77.90GB to 72.21GB. This memory efficiency is particularly valuable for resource-constrained environments. Most notably, energy consumption sees dramatic improvements with text-only training. LLaVA-1.6-7B's energy consumption decreases from 14.45Wh to 9.03Wh (37.5\% reduction), while Qwen2-VL-7B shows an even more significant reduction from 51.10Wh to 26.00Wh (49.1\% reduction). These efficiency gains, combined with the competitive performance shown in Table~\ref{tab:main_results}, suggest that text-only training offers a more sustainable and resource-efficient approach to improving VLM capabilities, particularly valuable in scenarios where computational resources are limited.

\section{Conclusion}

Our work demonstrates that text-only training can enhance fine-grained visual understanding in Vision-Language Models (VLMs), achieving comparable or even superior performance to image-text training while significantly reducing computational resources. Our results show that rich textual descriptions are effective in conveying visual concepts. This approach provides a sustainable, resource-efficient alternative for advancing VLMs, particularly valuable in scenarios where gathering image data is challenging or costly.

\section{Acknowledgements}
This research was supported by Brian Impact, a non-profit organization dedicated to advancing science and technology. 


\bigskip
\bibliography{aaai25}

\clearpage
\appendix

\section{Appendix A: Additional Details on Training}

\subsection{Models}

We experiment with two state-of-the-art VLMs: LLaVA-1.6~\cite{llava} and Qwen2-VL~\cite{Qwen2-VL}, both using the 7B parameter variant. LLaVA-1.6 builds on CLIP's~\cite{clip} vision encoder and Vicuna's~\cite{vicuna} language model, enhanced with visual reasoning capabilities through instruction tuning. It employs a projector structure to bridge the vision and language models, facilitating efficient multimodal learning.

Qwen2-VL extends the Qwen2 language model with visual capabilities through a modular design. It utilizes a vision transformer~\cite{vit} backbone followed by a multi-layer perceptron projector to align visual features with the language space. The model is pre-trained on a large-scale filtered dataset, incorporating both image-text pairs and interleaved image-text data.

Both models share similar architectural principles: a vision encoder for image processing, a projector for modality alignment, and a language model for text generation. They differ primarily in their pre-training approaches and the specific architectural choices in their projection layers. Following the official fine-tuning procedures of both models, we freeze the vision encoder parameters while fine-tuning the language model and projection layers.

\subsection{Training Setting}

We fine-tune the VLM models using both image-text and text-only training datasets. For both training approaches, we use the AdamW optimizer with a learning rate of 3e-5. Due to memory constraints and the model size, we set a batch size of 1 and employ gradient accumulation with 32 steps to achieve an effective batch size of 32. The training is conducted for a maximum of 5 epochs on a single NVIDIA A100 80GB GPU.

For the text-only training setting, we leverage the model's built-in capability to handle uni-modal inputs. The model processes these instances without any image placeholders or special tokens, maintaining its native architecture while handling text-only data. 
Given our small-data setting and independently designed evaluation dataset, we report the best performance achieved during the training process.

\section{Appendix B: Dataset Construction}

\subsection{Training Data Generation}

For training data generation, we designed two distinct prompts tailored to the characteristics of each domain. The cultural dataset prompt emphasizes comprehensive coverage of both tangible aspects (appearance, materials) and intangible elements (historical significance, traditional usage). The butterfly dataset prompt follows a more structured approach with specific sections covering physical characteristics, ecological factors, and behavioral patterns. Below are the detailed prompts and representative examples of the generated descriptions:

\begin{tcolorbox}[promptbox, title=Korean Cultural Dataset Generation]
\textbf{Prompt:}\\
Based on the provided image and {keyword}, provide a highly detailed and visually rich description document about the keyword. Do not mention this particular image directly, and use a general and typical manner of speech.

The description should include:
\begin{itemize}
   \item Visual characteristics and appearance
   \item Historical and cultural significance
   \item Traditional usage and context
   \item Regional variations and local characteristics
   \item Material composition and craftsmanship
\end{itemize}
\noindent\rule{\linewidth}{0.3pt}
\vspace{0.5em}
\textbf{Generated Example:}\\
Gat: The Traditional Korean Hat

The gat is a traditional Korean hat that has remained an iconic part of Korean cultural heritage. Worn primarily during the Joseon Dynasty (1392–1910), the gat was typically donned by men of noble or scholarly status and served both functional and symbolic purposes.

The gat consists of two main parts: The crown (Tongjeong), a cylindrical structure made of horsehair or bamboo designed to sit comfortably yet firmly on the head, and the brim (Yangtae), a wide circular extension crafted from thinly split bamboo and covered with a mixture of lacquer and black silk for its sleek appearance.

Creating a gat requires intricate craftsmanship combining skilled techniques and high-quality materials. Key materials include horsehair (Mawe) for the crown's lattice-like structure, bamboo (Daetjip) for both crown and brim framework, and silk and lacquer for the finishing. The craftsmanship is passed down through generations, with artisans (gatjiljang) requiring years of apprenticeship to master the techniques.

Different variations of the gat denoted hierarchy: Heungnip for scholars and officials symbolizing dignity, Baeknip for mourning periods representing purity, and Bangnip with broader brims for common use. While no longer worn daily, the gat remains significant in traditional ceremonies and cultural performances, inspiring modern designers who incorporate its elements into contemporary accessories [\ldots]
\end{tcolorbox}
\vspace{0.5em}

\begin{tcolorbox}[promptbox, title=Butterfly Dataset Generation]
\textbf{Prompt:}\\
Please write a detailed and thorough description of the butterfly species "{butterfly\_name}". Include the following sections:

\begin{enumerate}
   \item Visual Description
   \begin{itemize}
       \item Wing patterns, colors, unique physical traits
       \item Size variations and wing shapes
       \item Distinctive markings and differences from similar species
       \item Dorsal and ventral wing views
   \end{itemize}

   \item Habitat
   \begin{itemize}
       \item Typical regions and climates
       \item Vegetation types and geographic distribution
       \item Environmental preferences
   \end{itemize}

   \item Feeding Source
   \begin{itemize}
       \item Caterpillar and adult food sources
       \item Specific plant preferences
       \item Feeding behaviors and preferences
   \end{itemize}

   \item Breeding Cycle
   \begin{itemize}
       \item Breeding seasons and reproduction patterns
       \item Life cycle stage durations
       \item Environmental influences
   \end{itemize}

   \item Behavioral Patterns
   \begin{itemize}
       \item Flight characteristics
       \item Defense mechanisms
       \item Social behaviors and environmental interactions
   \end{itemize}
\end{enumerate}
\noindent\rule{\linewidth}{0.3pt}
\vspace{0.5em}
\textbf{Generated Example:}\\
Orange Oakleaf Butterfly

The Orange Oakleaf is a striking butterfly with distinctive wing patterns and coloration. Its dorsal wings display a rich red hue with intricate patterns of black and white, creating stunning contrast. The forewings feature black lines and spots, while hindwings have black edges with white or cream-colored spots. The ventral side shows more subdued brown and gray tones with complex patterns providing effective camouflage. The wingspan typically ranges from 5 to 7 centimeters, with slightly pointed forewings and rounded hindwings.

This species thrives in warm, humid tropical and subtropical environments, particularly in lowland rainforests with dense canopy. It inhabits areas from sea level up to 1,200 meters, preferring locations rich in flowering plants and fruiting trees. The caterpillars feed specifically on certain host plant families, while adults consume nectar from bright flowers and occasionally feed on tree sap, rotting fruit, or minerals from damp soil.

The breeding cycle occurs year-round in tropical regions, peaking during wet seasons. Males perform aerial displays to attract mates, after which females lay eggs on host plants. The complete life cycle spans 4-6 weeks: eggs hatch after a week, larvae develop for several weeks through multiple molts, and pupation lasts 10-14 days. The species exhibits erratic yet graceful flight patterns and shows strong territorial behavior while using its wing patterns both as warning coloration and camouflage against predators [\ldots]

\end{tcolorbox}

\subsection{Evaluation Data Generation}

For Type 2 questions in the butterfly dataset, which focus on testing ecological and biological understanding beyond visual recognition, we designed a specific prompt to generate multiple-choice questions:

\begin{tcolorbox}[promptbox, title=Butterfly Type 2 Question Generation]
\textbf{Prompt:}\\
Create a multiple-choice question about a butterfly species that focuses on its ecological or biological traits, such as habitat, feeding habits, lifecycle, behaviors, or role in the ecosystem. The question should test knowledge beyond its visual appearance and require an understanding of the species' biology or ecology.

Instructions:
\begin{itemize}
    \item Do not mention the butterfly's name in either the question or the answer options
    \item Focus on traits like habitat, diet, behavior, migration patterns, or lifecycle
    \item Ensure the question is specific and challenging, requiring knowledge of the species
\end{itemize}

Generate:
\begin{enumerate}
    \item A question that asks about a specific ecological or biological trait of the butterfly
    \item Four answer options labeled A), B), C), and D), with one correct and three plausible but incorrect options
\end{enumerate}

\noindent\rule{\linewidth}{0.3pt}
\vspace{0.5em}
\textbf{Generated Example:}\\
Q: Which feeding behavior is characteristic of this butterfly species in its adult stage?

A) Primarily feeds on tree sap and overripe fruit\\
B) Exclusively feeds on nectar from red flowers\\
C) Feeds on both nectar and minerals from damp soil\\
D) Only consumes water and dissolved sugars
\end{tcolorbox}

\section{Appendix C: Future Work}
While our current work demonstrates the effectiveness of text-only training in a small-data regime with 7B parameter models, several directions remain for future exploration:

\begin{itemize}
  \item Experiments with various types and scales of vision-language models (VLMs) and datasets
  \item Performance and error analysis on diverse vision-language tasks (e.g., visual description, captioning)
  \item Optimizing text description strategies and exploring hybrid approaches with limited image data
  \item Application to real-world scenarios with limited image data availability
\end{itemize}

These future directions could help establish text-only training as a practical solution for efficient VLM adaptation across various domains and resource settings.

\end{document}